\pdfoutput=1

\documentclass[11pt]{article}

\usepackage{ACL2023}

\usepackage{times}
\usepackage{latexsym}

\usepackage[T1]{fontenc}

\usepackage[utf8]{inputenc}

\usepackage{microtype}

\usepackage{dirtytalk}

\usepackage{algorithm}
\usepackage{algpseudocode}

\usepackage{inconsolata}
\usepackage{colortbl}
\usepackage{booktabs}
\usepackage{multirow}
\usepackage{subcaption}
\usepackage{amsmath}
\usepackage{pgfplots}
\usepackage{comment}

\newenvironment{itemize*}
    {\begin{itemize}%
      \setlength{\itemsep}{0pt}%
      \setlength{\parskip}{0pt}}%
    {\end{itemize}}

\newcommand{\tfidf}{\textsc{Tf$\times$Idf}}

\usepackage{booktabs}
\usepackage{siunitx} 
\usepackage{graphicx}

\title{Language Model Adaptation to Specialized Domains through Selective Masking based on Genre and Topical Characteristics} 

\author{
Anas Belfathi\textsuperscript{} \quad
Ygor Gallina\textsuperscript{} \quad
Nicolas Hernandez\textsuperscript{} \quad
Richard Dufour\textsuperscript{} \quad
Laura Monceaux\textsuperscript{} \\
\textsuperscript{}LS2N, UMR CNRS 6004, Nantes Université \\
\texttt{\{firstname.lastname\}@univ-nantes.fr}
}

\pgfplotsset{compat=1.18} 
\begin{document}
\maketitle

\begin{abstract}
Recent advances in pre-trained language modeling have facilitated significant progress across various natural language processing (NLP) tasks. Word masking during model training constitutes a pivotal component of language modeling in architectures like BERT. However, the prevalent method of word masking relies on random selection, potentially disregarding domain-specific linguistic attributes. In this article, we introduce an innovative masking approach leveraging genre and topicality information to tailor language models to specialized domains. Our method incorporates a ranking process that prioritizes words based on their significance, subsequently guiding the masking procedure. Experiments conducted using continual pre-training within the legal domain have underscored the efficacy of our approach on the LegalGLUE benchmark in the English language. Pre-trained language models and code are freely available for use.





\end{abstract}

\section{Introduction}

Large-scale, pre-trained language models (PLMs) have become indispensable in modeling human language, significantly advancing performance across diverse Natural Language Processing (NLP) tasks~\cite{bao2020unilmv2, guu2020realm, zhang2022opt}. Among these architectures, masked language models (MLMs) like BERT~\cite{devlin-etal-2019-bert} are prominent, where 
tokens within a sequence are intentionally masked during training, compelling the model to predict these tokens based on surrounding context. This method enables the model to grasp intricate semantic relationships and syntactic structures inherent in natural language. However, training MLMs from scratch demands substantial resources in terms of data and computing power.

In the context of specialized domains,  adaptation through continual pre-training remains the conventional approach, drawing upon domain-specific data to refine pre-trained models~\cite{chalkidis-etal-2020-legal,wu2021pretrained, ke2022continual,labrak-etal-2023-drbert}. This process invariably entails token masking, where critical factors such as the masking ratio and token selection play pivotal roles. Prior research efforts~\cite{sun_ernie_2019, joshi-etal-2020-spanbert, levine2020pmimasking, li-zhao-2021-pre} have delved into selective information fine-tuning, encompassing words, tokens, and spans.

In this paper, we introduce an original masking approach that harnesses genre and topicality information to tailor MLMs to specialized domains. Our method integrates a ranking process with meta-discourse and \tfidf{} scoring to prioritize tokens based on contextual significance, guiding the masking procedure. By systematically identifying and masking tokens crucial to domain-specific contexts, we compel the model to adapt to the understanding and prediction of essential domain-specific words. To illustrate the effectiveness of our strategies, we conduct experiments 
on the continual pre-training (CPT) of BERT models towards the legal domain, comparing various token masking strategies.


Our contributions include:

\begin{itemize*}
    \item We propose original masking strategies based on word selection (meta-discourse and \tfidf{}) for language model training.
    \item We develop a systematic approach for incorporating selected words effectively during the training process.
    \item We release open-source models and code designed for adaptable training, facilitating MLM pre-training for specific domains based on our approach\footnote{\href{https://github.com/ygorg/legal-masking}{github.com/ygorg/legal-masking}}.
\end{itemize*}

\section{Related work}



The classical masking strategy in BERT~\cite{devlin-etal-2019-bert} is the masking of 15\% of tokens within a given sentence. The approach involves the random replacement (10\% chance), preservation (10\%), or substitution with the special \texttt{[MASK]} token (80\%) of selected tokens. The model's objective is to accurately predict the original tokens.

In efforts to enrich the representation capabilities of MLMs, ERNIE~\cite{sun2019ernie} and SpanBERT~\cite{joshi-etal-2020-spanbert} have refined the classical strategy of random token masking employed by BERT. They introduced methods emphasizing the masking of entire words and spans of text, respectively, albeit still in a random manner. These approaches have demonstrated improved performance in certain domain-specific tasks.

From a different perspective, recent studies have explored dynamically altering the content masked during the training process. \citet{yang_learning_2023} introduced a time-variant masking strategy, departing from static methods that maintain consistent content throughout training. They noted that, at certain training stages, models cease to learn from specific types of words, discerned through part-of-speech tags and the model's error rates using loss. Similarly, \citet{althammer_linguistically_2021} applied masking to words within noun chunks, manipulating the predictability probability of such tokens.

To our knowledge, our work is the first to propose and investigate the impact of masking strategies employing semantically important words automatically selected for a specialized domain.


\section{Selective Masking}

While the original BERT approach employed random word selection~\cite{devlin-etal-2019-bert}, our method selectively masks words based on their significance to a specific text genre or their topical salience 
within a document. 
We adopt a two-step approach: firstly, assigning a "genre specificity score" and "topical salience score" to each word from a domain-specific corpus (Section~\ref{ssec:importance_scoring}). Subsequently, we use these ranked lists to determine which words to mask (Section~\ref{s:mask_strat}).  




\subsection{Word Weighting Approaches}
\label{ssec:importance_scoring}



We propose two automatic word weighting approaches computed from a set of domain-specific documents.
The first approach, the \textit{topicality score} (\texttt{\tfidf{}}), quantifies the thematic relevance of a word to a given document. We employ the well-established \tfidf{}~\cite{Jones_1972} score, which evaluates a word's topical salience by comparing its frequency in a document to its occurrence across multiple documents. 


The second approach, the \textit{specificity score for a text genre} (\texttt{MetaDis}), assesses the extent to which a word is characteristic of a particular text genre. A genre of documents is characterized by a common structure~\cite{biber_2019_register}, often described by words or expressions known as \textit{meta-discourse}~\cite{hyland1998persuasion}. For instance, in the legal genre of jurisprudence, this lexicon includes terms used to describe facts, present arguments, reason, or reach a final decision. While \citet{hernandez_2003_automatic} utilized the inverse document frequency score to assess specificity, this measure overlooks the distribution of occurrences within documents. We assume that a meta-discourse marker occurs in a consistent proportion across documents within the same genre. To capture such properties and calculate a meta-discourse score, we propose the formula depicted in Equation~\ref{eq:meta-discourse}:

\vspace{-.5cm}

\begin{equation}
s_t = \frac{df_t}{tf_t} * \left( 1-\frac{std(dtf_t)}{max(dtf_t)} \right) * \frac{df_t}{N}
\label{eq:meta-discourse}
\end{equation}

Here, $df_t$ and $tf_t$ represent the document frequency and term frequency of a specific word, $t$, respectively, while $N$ signifies the total number of documents in the corpus. $dtf_t$ lists the number of occurrences per document for a given word. The first term in the equation weights a word based on its occurrence across distinct documents and its total number of occurrences. A word that appears a few times per document receives a higher score. The second term accounts for words with consistent occurrences across documents, reflected by a low standard deviation. Finally, the third term emphasizes words that appear in multiple documents, contributing to their overall score.

\subsection{Word Selection Strategy}
\label{s:mask_strat}

We propose two strategies for selecting words to mask from a weighted list of words obtained previously. Our first method, \texttt{TopN}, selects the top 15\% of words with the highest scores.
The second method, \texttt{Rand}, aims to enhance model robustness by avoiding systematic masking of the same words. This method introduces a level of weighted randomness, similar to the dynamic masking approach used in RoBERTa ~\cite{liu_2019_roberta}.
In practice, we randomly sample words (without replacement) based on the distribution of computed scores (refer to Algorithm~\ref{alg:mask_choice} in Appendix~\ref{sec:masked_analysis}).
%


%

\section{Experimental setup}

\vspace{-.2cm}
We utilize BERT~\cite{devlin-etal-2019-bert} and LegalBERT~\cite{chalkidis_legal-bert_2020} as our pre-trained base models.
The effectiveness of our masking strategies is assessed through continual pre-training on these models, focusing on legal domain adaptation.
To facilitate this process, we introduce a pre-training dataset collected specifically for this purpose, which is also utilized to select masking words (Section~\ref{s:corpus}).
We delineate the evaluation tasks (Section~\ref{s:eval}) and conclude with the experimental specifics (Section~\ref{s:exp_details}).

\subsection{Pre-training Corpus}
\label{s:corpus}

\begin{table}[ht!]
\centering
\small
\definecolor{grayshade}{gray}{0.9} 
\begin{tabular}{l|r|rl}
\hline
\rowcolor{white} 
\textbf{Sub-Corpus} & \textbf{\# Doc} & \textbf{\# Tokens} & \\
\hline
\rowcolor{grayshade}
EU Case Law & 29.8K & 178.5M & 29\% \\
\rowcolor{white}
ECtHR Case Law & 12.5K & 78.5M & 13\%\phantom{.0} \\
\rowcolor{grayshade}
U.S. Case Law & 104.7K & 235.5M & 39\%\phantom{.0} \\
\rowcolor{white}
Indian Case Law & 34.8K & 111.6M & 19\% \\ 
\hline
\rowcolor{white} 
\textbf{Total} & \textbf{181.8K} & \textbf{604.1M} & \textbf{100\%} \\
\hline
\end{tabular}
\caption{Details of the 4Gb dataset for CPT.}
\label{tab:dataset_desc}
\end{table}

\vspace{-.3cm}

For continual pre-training and word masking selection, we chose to focus on the legal domain by utilizing a subset of the LexFiles corpus~\cite{chalkidis_lexfiles_2023} representative of the LexGLUE~\cite{chalkidis_lexglue_2022} benchmark. The documents were selected to offer a balanced and diverse collection, encompassing the linguistic nuances 
(see Table~\ref{tab:dataset_desc}). 


\subsection{Evaluation Tasks}
\label{s:eval}

We assess the performance of our model using LexGLUE~\cite{chalkidis_lexglue_2022}, a benchmarking framework designed specifically for the legal domain. LexGLUE encompasses a diverse array of legal tasks sourced from European, United States, and Canadian legal systems. These tasks entail multi-class and multi-label classification at the document level, with a dozen labels. Such a setup provides a rigorous test for our approach's ability to excel across a spectrum of complex tasks.

\subsection{Experimental Details}
\label{s:exp_details}

For continual pre-training, we conducted sessions totaling over 20 hours using 16 V100 GPUs on the Jean Zay supercomputer. We adopted a batch size of 16 and configured gradient accumulation steps to 16, resulting in an effective batch size of 4096, following the methodology outlined by~\citet{labrak-etal-2023-drbert}. To ascertain task performance scores, we computed the average of scores from three independent runs. Results are presented in terms of micro ({\( \mu F_1 \)}) and macro (m-F1) F-measure.

\section{Results and Discussions}

\begin{table*}[!htbp]
\centering

\resizebox{\textwidth}{!}{
\begin{tabular}{
  l
  l
  l
  r
  r
  r
  r
  r
  r
  r
  r
  r
  r
}
\toprule
\textbf{Method} & \multicolumn{2}{c}{\textbf{ECtHR (A)}} & \multicolumn{2}{c}{\textbf{ECtHR (B)}} & \multicolumn{2}{c}{\textbf{SCOTUS}} & \multicolumn{2}{c}{\textbf{EUR-LEX}} & \multicolumn{2}{c}{\textbf{LEDGAR}} & \multicolumn{2}{c}{\textbf{UNFAIR-ToS}}\\
& {\( \mu F_1 \)} & {m-F1} & {\( \mu F_1 \)} & {m-F1} & {\( \mu F_1 \)} & {m-F1} & {\( \mu F_1 \)} & {m-F1} & {\( \mu F_1 \)} & {m-F1} & {\( \mu F_1 \)} & {m-F1} \\
\midrule
\vspace{.1cm}%
BERT & 62.12 & 52.66 & 69.59 & 60.39 & 69.61 & 58.65 & 71.70 & 54.87 & \underline{87.85} & \textbf{82.30} & \textbf{95.66} & 80.97 \\
\vspace{.1cm}%
+ CPT (baseline) & \underline{63.12} & 54.13 & 71.06 & \textbf{64.69} & \underline{70.57} & \textbf{60.38} & \underline{71.86} & 56.18 & \textbf{87.90} & 82.02 & 95.56 & \underline{81.46} \\
+ MetaDis - Rand & \cellcolor{green!15} 62.55 & \cellcolor{green!50} 54.88 & \cellcolor{green!15} 70.45 & \cellcolor{green!15} 63.10 & \cellcolor{green!15} 70.26 &  \cellcolor{green!15}59.12 & \cellcolor{green!15} 71.66 & \cellcolor{green!15} 56.00 & \cellcolor{green!15} 87.68 & \cellcolor{green!50} \underline{82.25} & \cellcolor{green!15} 95.38 & \cellcolor{green!15} 79.18 \\
\vspace{.1cm}%
+ MetaDis - TopN & \cellcolor{green!15} 62.17 & \cellcolor{green!15} 53.35 & \cellcolor{green!15} 70.29 & \cellcolor{green!15} 62.29 & \cellcolor{green!15} 69.92 & \cellcolor{green!15} \underline{60.08} & \cellcolor{green!15} 71.67 & \cellcolor{green!50} \underline{56.95} & \cellcolor{green!15} 87.78 & \cellcolor{green!50} 82.11 & \cellcolor{green!50} \underline{95.57} & \cellcolor{green!50} \textbf{81.51} \\

+ \tfidf{} - Rand & \cellcolor{orange!50} \textbf{63.36} & \cellcolor{orange!50} \textbf{56.60} & \cellcolor{orange!50} \underline{71.32} & \cellcolor{orange!15}   \underline{64.58} & \cellcolor{orange!15}   69.69 & \cellcolor{orange!15}   59.10 & \cellcolor{orange!50} \textbf{71.93} & \cellcolor{orange!15}   55.82 & \cellcolor{orange!15}   87.69 & \cellcolor{orange!50} 82.11 & \cellcolor{orange!15}   95.50 & \cellcolor{orange!15}   78.63\\

+ \tfidf{} - TopN & \cellcolor{orange!15}  62.66 & \cellcolor{orange!50}  \underline{56.46} & \cellcolor{orange!50}  \textbf{71.50} & \cellcolor{orange!15}  63.58 & \cellcolor{orange!50}  \textbf{70.71} & \cellcolor{orange!15}  60.06 & \cellcolor{orange!15}  71.73 & \cellcolor{orange!50}  \textbf{57.73} & \cellcolor{orange!15}  87.67 & \cellcolor{orange!15}  81.89 & \cellcolor{orange!15}  95.49 & \cellcolor{orange!15}  79.45 \\
\midrule

\vspace{.1cm}
LegalBERT & 63.41 & 53.19 &  72.10 & 63.68 & 73.61 & 61.50 & 71.93 & \underline{55.47} & 87.91 & 81.67 & \underline{95.81} & \underline{81.27} \\
\vspace{.1cm}%
+ CPT (baseline) & \underline{63.64} & \textbf{58.73} & 72.60 & 64.95 & \textbf{74.64} & \textbf{63.13} & \underline{72.01} & 55.12 & \textbf{88.41} & \textbf{82.92} & \textbf{95.82} & 79.70 \\
+ MetaDis - Rand & \cellcolor{green!15} 63.39 & \cellcolor{green!15} 56.39 & \cellcolor{green!50} \underline{73.08} & \cellcolor{green!50} 65.76 & \cellcolor{green!15} 74.21 & \cellcolor{green!15} 62.97 & \cellcolor{green!50} \textbf{72.03} & \cellcolor{green!15} 54.76 & \cellcolor{green!15} \underline{88.38} & \cellcolor{green!15} 82.58 & \cellcolor{green!15} 95.20 & \cellcolor{green!50} 80.26 \\
\vspace{.1cm}
+ MetaDis - TopN & \cellcolor{green!50} \textbf{64.07} & \cellcolor{green!15} \underline{58.56} & \cellcolor{green!15} 72.53 & \cellcolor{green!50} \textbf{66.83} & \cellcolor{green!15} 73.88 & \cellcolor{green!15} 62.57 & \cellcolor{green!15} 71.96 & \cellcolor{green!15} 55.01 & \cellcolor{green!15} 88.32 & \cellcolor{green!15} 82.16 & \cellcolor{green!15} 94.80 & \cellcolor{green!15} 73.67\\

+ \tfidf{} - Rand & \cellcolor{orange!15}  63.38 & \cellcolor{orange!15}  56.78 & \cellcolor{orange!15}  72.21 & \cellcolor{orange!50}  65.67 & \cellcolor{orange!15}  73.71 & \cellcolor{orange!15}  62.85 & \cellcolor{orange!15}  71.78 & \cellcolor{orange!50}  \textbf{55.82} & \cellcolor{orange!15}  88.19 & \cellcolor{orange!15}  82.36 & \cellcolor{orange!15}  95.80 & \cellcolor{orange!50}  \textbf{82.12 }\\

+ \tfidf{} - TopN & \cellcolor{orange!15}  62.89 & \cellcolor{orange!15}  53.58 & \cellcolor{orange!50}  \textbf{73.26} & \cellcolor{orange!50}   \underline{65.86} & \cellcolor{orange!15}  \underline{74.38} & \cellcolor{orange!15}  \underline{63.10} & \cellcolor{orange!15}  71.90 & \cellcolor{orange!15}  55.08 & \cellcolor{orange!15}  88.27 & \cellcolor{orange!15}  \underline{82.65} & \cellcolor{orange!15}  95.59 & \cellcolor{orange!50}  81.20\\
\bottomrule
\end{tabular}
}
\caption{Comparative analysis of BERT and LegalBERT models' performance using continual pre-training with selective masking on LexGLUE benchmark tasks. Best performing models are indicated in bold, second-best results are underlined. \colorbox{green!15}{Green} and \colorbox{orange!15}{Orange} respectively depict MetaDis and \tfidf{} scores. Darker colors highlight improvement over the CPT (baseline).}
\label{tab:lexglue_res}
\end{table*}

Results are detailed in Table~\ref{tab:lexglue_res}, each column referring to a task of the LexGLUE benchmark.

\paragraph{Efficacy of Continual Pre-training} Our study reveals the effectiveness of continual pre-training across all tasks and experiments for both BERT and LegalBERT models. Specifically, the micro-F1 ({\( \mu F_1 \)}) score for the baseline BERT+CPT demonstrates a notable improvement from 60.39 to 64.69 in the ECtHR (B) task, while the baseline LegalBERT-CPT shows substantial enhancements in the EUR-LEX task. These improvements, even without modifying the masking strategy, suggest that our collected corpus contains new domain-specific characteristics that enrich the underlying knowledge of language models.

\paragraph{Classical vs. Selective Masking} To evaluate the efficacy of our masking approaches, we compared them with the classical BERT random masking in a continual pre-training setup (+CPT (baseline)).
The results indicate that our models improve over the baseline across all tasks, irrespective of the masking strategy employed.
With BERT, notable improvements were observed in the ECtHR(A) and LEDGAR tasks, with scores of 54.88 and 82.25, respectively.
This can be attributed to our models' enhanced ability to leverage Genre (MetaDis) and thematic (\tfidf{}) information inherent in the legal domain.
Conversely, for the LegalBERT models, improvements were noted in the ECtHR(B) and UNFAIR-ToS tasks under both scoring methods.
This underscores the benefits of selectively masking words, especially with already adapted models.
Moreover, our approach demonstrated superior results on the SCOTUS task compared to hierarchical approaches mentioned in~\citet{chalkidis_lexglue_2022}, employing a streamlined and less complex model structure. 
This highlights the importance of choosing masking techniques that focus on domain-specific language features, avoiding complex models with extra parameters or longer training times.

\paragraph{MetaDiscourse vs. \tfidf{}} When comparing genre (+MetaDis) and thematic scoring (+\tfidf{}) masking strategies on BERT and LegalBERT models, we observed distinct patterns. For the BERT models, meta-discourse demonstrated its effectiveness in tasks where genre-specific language features play a significant role, such as in the ECtHR (A), LEDGAR, and UNFAIR-ToS tasks. In contrast, \tfidf{} showed its strengths in tasks that emphasize thematic relevance, such as in the ECtHR (B), EURLEX, and SCOTUS tasks.

For the LegalBERT models, both strategies displayed similar performance in the ECtHR (B) and UNFAIR-ToS tasks.
However, meta-discourse proved to be more effective in the ECtHR (A) task, while \tfidf{} demonstrated better results for EURLEX.
These findings suggest that thematic relevance is generally crucial for the EURLEX task regardless of the model's starting point, indicating that the thematic scoring aligns well with the task's nature.
Conversely, genre considerations (MetaDis) are particularly beneficial for the ECtHR (A) task, emphasizing the importance of structural and stylistic language features in legal texts.

\paragraph{Rand vs. TopN} The experimental analysis of BERT models revealed that the random weighting strategy (Rand) was successful in achieving higher performances in the ECtHR (A) and LEDGAR tasks for both scoring methods.
On the other hand, the TopN strategy showed improvements in the EURLEX task with these scoring techniques.
Notably, the TopN method also demonstrated higher performance in the ECtHR (B) and SCOTUS tasks when using \tfidf{}, indicating its effectiveness in situations where topicality is crucial.

Regarding the LegalBERT models with meta-discourse, the Rand strategy was effective in enhancing performance in the UNFAIR-ToS, ECtHR (B), and EURLEX tasks. Meanwhile, the TopN approach made significant strides in both ECtHR (A) and (B), emphasizing the importance of focusing on the most pertinent words.
When using \tfidf{} with LegalBERT, improvements were observed in the ECtHR (B) and UNFAIR-ToS tasks under both strategies. Interestingly, the Rand approach achieved the best results for the EURLEX task.
The consistent improvements across various tasks confirm the interest in selective masking, though it might require customization for the task.


\section{Conclusion and Future Work}

Our research provides conclusive evidence that our proposed automatic selective masking strategies integrating genre and topical characteristics played a crucial role in refining the models' focus when adapted to a specialized domain. We observed improvements across all tasks in the LexGLUE benchmark focusing on the legal domain. Notably, both the BERT and LegalBERT models demonstrated important improvements in the ECtHR and EUR-LEX tasks.
While our results are encouraging, several avenues for further research emerge, including the impact of our approach on other domains, such as the clinical and scientific domains, to ensure the generalizability of our approach.
Furthermore, it is crucial to tackle the obstacles presented by multi-language models.

\section*{Ethical considerations}

With respect to the potential risks and biases inherent in language models trained on legal datasets, legal corpora may comprise texts of varying quality and representativeness.
The utilization of models such as BERT trained on legal texts could potentially introduce biases pertaining to fairness, the use of gendered language, the representation of minority groups, and the dynamic nature of legal standards over time.
It is imperative that these biases are thoroughly evaluated and mitigated to ensure equitable performance across different demographics and to maintain currency with evolving legal norms.

\section*{Limitations}

Our work, though offering valuable insights into the application of continual pre-training and selective masking techniques for language models in the legal domain, is not without its limitations.
Specifically, the current study concentrates solely on the BERT architecture, which restricts our ability to investigate a broader range of language models that may exhibit distinct behaviors and sensitivities to our pre-training and masking strategies. Future studies should explore other models, such as DrBERT~\cite{labrak-etal-2023-drbert} and RoBERTa~\cite{liu_2019_roberta}, to provide a more comprehensive understanding of the effects of our approach.
Additionally, our study lacks a direct comparison with a pre-trained model developed using selective masking from scratch. Such a comparison would serve as a valuable reference point for assessing the incremental benefits of our method. 
Finally, we acknowledge that further hyperparameter tuning may lead to enhanced model performance.


\section*{Acknowledgements}
This work was granted access to the HPC resources of IDRIS under the allocation 2023-AD011014882 made by GENCI.

This research was funded, in whole or in part, by l'Agence Nationale de la Recherche (ANR), project NR-22-CE38-0004.

\bibliography{custom}
\bibliographystyle{acl_natbib}

\appendix

\section{Continual Pre-training parameters}
\label{sec:appendix}



Before training, samples were randomly shuffled 3 times using the same seed.

We train each model using the \texttt{transformers} python library for 10 epochs which represents 4453 steps for bert and 4396 steps for legal-bert. The difference in the number of steps arises from the difference of tokenisation between legal-bert and bert, which results in a different number of training sequence.


In total we estimate the total computation time to $\simeq$ 4,100h which breaks down into 3,200h of training time, 380h for evaluating the models and 520h for development.


\section{Selection and Analysis of masked words}
\label{sec:masked_analysis}

We detail in Algorithm~\ref{alg:mask_choice} the process of selecting the words to mask.
The \tfidf{} was computed using the \texttt{scikit-learn} python package.

To gain more insight on the difference of selected words by the two importance scores, we show in the Table~\ref{tab:ex_masked_words} the 50 most masked words for $\simeq$10\% of the training corpus.


\begin{algorithm}
\caption{Explicit masking}
\begin{algorithmic}[1]

\Function{Mask}{$tokens$}
    \State $\mathcal{M} \leftarrow \{\}$
    \State $W \leftarrow WholeWords(tokens)$
    \State $S \leftarrow ScoreSequence(W)$

    \While{$|\mathcal{M}| < 0.15 * |tokens|$}
        \State $i \leftarrow \text{Sample}(S)$\footnotemark
        \State Remove $W[i]$ and $S[i]$

        \If{$|\mathcal{M}| + |W[i]| \leq 0.15 * |tokens|$}
            \State $\mathcal{M} \leftarrow \mathcal{M} + w$
        \EndIf
    \EndWhile
    \State \Return {$\mathcal{M}$}
\EndFunction
\end{algorithmic}
\label{alg:mask_choice}
\end{algorithm}
\footnotetext{Use the Max function for \texttt{TopN} method}

\begin{table*}[!htb]
    \centering
    \begin{tabular}{p{15.5cm}}
        \textbf{\tfidf} \\
        applicant, court, 2007, extradition, prosecutor, meshchanskiy, russian, dzhurayev, moscow, uzbekistan, tashkent, district, custody, government, convention, article, office, decision, §, detention, russia, ccp, 4, preventive, v, minsk, federation, 2, application, uzbek, proceedings, 1, 5, criminal, procedure, january, 38124, case, 29, merits, may, dismissed, law, rakhimovskiy, 466, request, decided, sobir, arrest, provisions \\
        \textbf{MetaDis} \\
        general, application, decision, january, september, decided, august, 4, 28, 9, 3, issued, request, rules, dismissed, 23, 29, indicated, basis, ordered, european, apply, be, 24, 17, date, 5, 30, held, final, december, 26, 6, 11, mentioned, applied, specified, 12, february, placed, 2, whether, remain, first, to, deliberated, represented, constitute, case, article 
    \end{tabular}
    \caption{50 most masked words using \tfidf{} and Metadiscourse scoring ordered by frequency.}
    \label{tab:ex_masked_words}
\end{table*}

\end{document}